\documentclass[journal]{IEEEtran}
\usepackage{amsmath,graphicx}
\usepackage{float}
\usepackage[caption = false]{subfig}
\usepackage{algorithm}
\usepackage{algpseudocode}
\usepackage{textcomp}
\usepackage{url}
\algtext*{EndWhile}
\algtext*{EndFor}
\algtext*{EndIf}
\DeclareMathOperator*{\argmax}{\arg\!\max}

\ifCLASSINFOpdf
\else
\fi

\begin{document}

%
\title{An End-to-End Neural Network for Polyphonic Piano Music Transcription}
%
%
%

\author{Siddharth~Sigtia,
        Emmanouil~Benetos,
        and~Simon~Dixon
\thanks{The authors are with the Centre for Digital Music, School of Electronic Engineering and Computer Science, Queen Mary University of London, E1 4NS, London, U.K. }
\thanks{EB is supported by a Royal Academy of Engineering Research Fellowship (grant no. RF/128).}
\thanks{Email: \{s.s.sigtia,emmanouil.benetos,s.e.dixon\}@qmul.ac.uk}
}        
\maketitle

\begin{abstract}

We present a supervised neural network model for polyphonic piano music transcription. The architecture of the proposed model is analogous to speech recognition systems and comprises an \emph{acoustic model} and a \emph{music language model}. The acoustic model is a neural network used for estimating the probabilities of pitches in a frame of audio. The language model is a recurrent neural network that models the correlations between pitch combinations over time. The proposed model is general and can be used to transcribe polyphonic music without imposing any constraints on the polyphony. The acoustic and language model predictions are combined using a probabilistic graphical model. Inference over the output variables is performed using the beam search algorithm. We perform two sets of experiments. We investigate various neural network architectures for the acoustic models and also investigate the effect of combining acoustic and music language model predictions using the proposed architecture. We compare 
performance of the neural network based acoustic models with two popular unsupervised acoustic models. Results show that convolutional neural network acoustic models yields the best performance across all evaluation metrics. We also observe improved performance with the application of the music language models. Finally, we present an efficient variant of beam search that improves performance and reduces run-times by an order of magnitude, making the model suitable for real-time applications.


\end{abstract}


\begin{center} \bfseries EDICS Category: AUD-MSP, AUD-MIR, MLR-DEEP \end{center}
%
\IEEEpeerreviewmaketitle

\section{Introduction}

\IEEEPARstart{A}{utomatic} Music Transcription (AMT) is a fundamental problem in Music Information Retrieval (MIR). AMT aims to generate a symbolic, score-like transcription, given a \emph{polyphonic} acoustic signal. Music transcription is considered to be a difficult problem even by human experts and current music transcription systems fail to match human performance \cite{klapuri2007signal}. Polyphonic AMT is a difficult problem because concurrently sounding notes from one or more instruments cause a complex interaction and overlap of harmonics in the acoustic signal. Variability in the input signal also depends on the specific type of instrument being used. Additionally, AMT systems with unconstrained polyphony have a combinatorially very large output space, which further complicates the modeling problem. 
Typically, variability in the input signal is captured by models that aim to learn the timbral properties of the instrument being transcribed \cite{berg2014unsupervised,benetos2012shift}, while the issues relating to a large output space are dealt with by constraining the models to have a maximum polyphony \cite{klapuri2003multiple,emiya2008automatic}. 

The majority of current AMT systems are based on the principle of describing the input magnitude spectrogram as a weighted combination of basis spectra corresponding to pitches. The basis spectra can be estimated by various techniques such as non-negative matrix factorisation (NMF) and sparse decomposition. Unsupervised NMF approaches \cite{smaragdis2003non,abdallah2004polyphonic} aim to learn a dictionary of pitch spectra from the training examples. However purely unsupervised approaches can often lead to bases that do not correspond to musical pitches, therefore causing issues with interpreting the results at test time. These issues with unsupervised spectrogram factorisation methods are addressed by incorporating harmonic constraints in the training algorithm \cite{vincent2010adaptive,bertin2010enforcing}. 
Spectrogram factorisation based techniques were extended with the introduction of probabilistic latent component analysis (PLCA) \cite{smaragdis2006probabilistic}. 
PLCA aims to fit a latent variable probabilistic model to normalised spectrograms. PLCA based models are easy to train with the expectation-maximisation (EM) algorithm and have been extended and applied extensively to AMT problems \cite{grindlay2010probabilistic,benetos2012shift}.

As an alternative to spectrogram factorisation techniques, there has been considerable interest in discriminative approaches to AMT. Discriminative approaches aim to directly classify features extracted from frames of audio to the output pitches. This approach has the advantage that instead of constructing instrument specific generative models, complex classifiers can be trained using large amounts of training data to capture the variability in the inputs. When using discriminative approaches, the performance of the classifiers is dependent on the features extracted from the signal. Recently, neural networks have been applied to raw data or low level representations to jointly learn the features and classifiers for a task \cite{lecun2015deep}. Over the years there have been many experiments that evaluate discriminative approaches for AMT. Poliner and Ellis \cite{poliner2007discriminative} use support vector machines (SVMs) to classify normalised magnitude spectra. Nam et. al. \cite{nam2011classification} 
superimpose an SVM on top of a deep belief network (DBN) in order to learn the features for an AMT task. Similarly, a bi-directional recurrent neural network (RNN) is applied to magnitude spectrograms for polyphonic transcription in \cite{bock2012polyphonic}. 

In large vocabulary speech recognition systems, the information contained in the acoustic signal alone is often not sufficient to resolve ambiguities between possible outputs. A language model is used to provide a prior probability of the current word given the previous words in a sentence. Statistical language models are essential for large vocabulary speech recognition \cite{rabiner1993fundamentals}. Similarly to speech, musical sequences exhibit temporal structure. In addition to an accurate acoustic model, a model that captures the temporal structure of music or a music language model (MLM), can potentially help improve the performance of AMT systems. Unlike speech, language models are not common in most AMT models due to the challenging problem of modelling the combinatorially large output space of polyphonic music. 
Typically, the outputs of the acoustic models are processed by pitch specific, two-state hidden Markov models (HMMs) that enforce smoothing and duration constraints on the output pitches \cite{benetos2012shift,poliner2007discriminative}. However, extending this to modelling the high-dimensional outputs of a polyphonic AMT system has proved to be challenging, although there are some studies that explore this idea. A dynamic Bayesian network is used in \cite{raczynski2013dynamic}, to estimate prior probabilities of note combinations in an NMF based transcription framework. 
Similarly in \cite{sigtiarnn}, a recurrent neural network (RNN) based MLM is used to estimate prior probabilities of note sequences, alongside a PLCA acoustic model. A sequence transduction framework is proposed in \cite{boulanger2013high}, where the acoustic and language models are combined in a single RNN.   

The ideas presented in this paper are extensions of the preliminary experiments in \cite{sigtia2014hybrid}. We propose an end-to-end architecture for jointly training both the acoustic and the language models for an AMT task. We evaluate the performance of the proposed model on a dataset of polyphonic piano music. We train neural network acoustic models to identify the pitches in a frame of audio. The discriminative classifiers can in theory be trained on complex mixtures of instrument sources, without having to account for each instrument separately. The neural network classifiers can be directly applied to the time-frequency representation, eliminating the need for a separate feature extraction stage. In addition to the deep feed-forward neural network (DNN) and RNN architectures in \cite{sigtia2014hybrid}, we explore using convolutional neural nets (ConvNets) as acoustic models. ConvNets were initially proposed as classifiers for object recognition in computer vision, but have found increasing application 
in speech recognition \cite{abdel2012applying,abdel2013exploring}. Although ConvNets have been applied to some problems in MIR \cite{schluter2014improved,humphrey2012rethinking}, they remain unexplored for transcription tasks. We also include comparisons with two state-of-the-art spectrogram factorisation based acoustic models \cite{benetos2012shift,vincent2010adaptive} that are popular in AMT literature. As mentioned before, the high dimensional outputs of the acoustic model pose a challenging problem for language modelling. We propose using RNNs as an alternative to state space models like factorial HMMs \cite{vincent2004music} and dynamic Bayesian networks \cite{raczynski2013dynamic}, for modeling the temporal structure of notes in music. RNN based language models were first used alongside a PLCA acoustic model in \cite{sigtiarnn}. However, in that setup, the language model is used to iteratively refine the predictions in a feedback loop resulting in a non-causal and theoretically unsatisfactory model. In 
the hybrid framework, \emph{approximate} inference over the output variables is performed using beam search. However beam search can be computationally expensive when used to decode long temporal sequences. We apply the efficient hashed beam search algorithm proposed in \cite{sigtiachords} for inference. The new inference algorithm reduces decoding time by an order of magnitude and makes the proposed model suitable for real-time applications. Our results show that convolutional neural network acoustic models outperform the remaining acoustic models over a number of evaluation metrics.  We also observe improved performance with the application of the music language models.

The rest of the paper is organised as follows: Section \ref{section:background} describes the neural network models used in the experiment, Section \ref{section:methods} discusses the proposed model and the inference algorithm, Section \ref{section:evaluation} details model evaluation and experimental results. Discussion, future work and conclusions are presented in Section \ref{section:discussion}.


\begin{figure*}[ht]
\subfloat[DNN]{\includegraphics[width=0.1\textwidth,height=6cm]{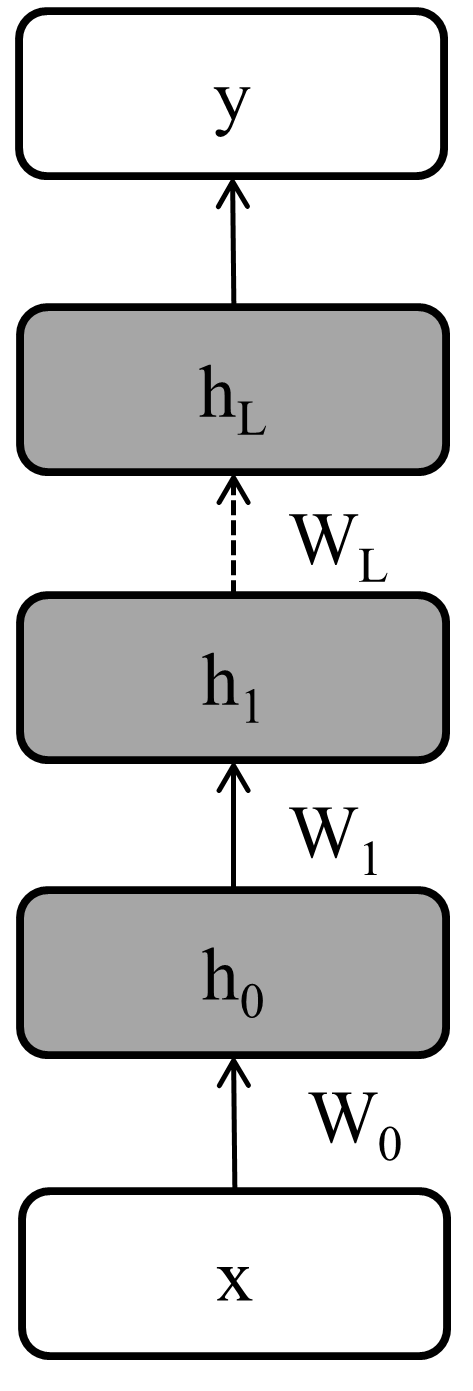}\label{DNN_fig}}
\hspace{20px}
\subfloat[RNN]{\includegraphics[width=0.15\textwidth,height=6cm]{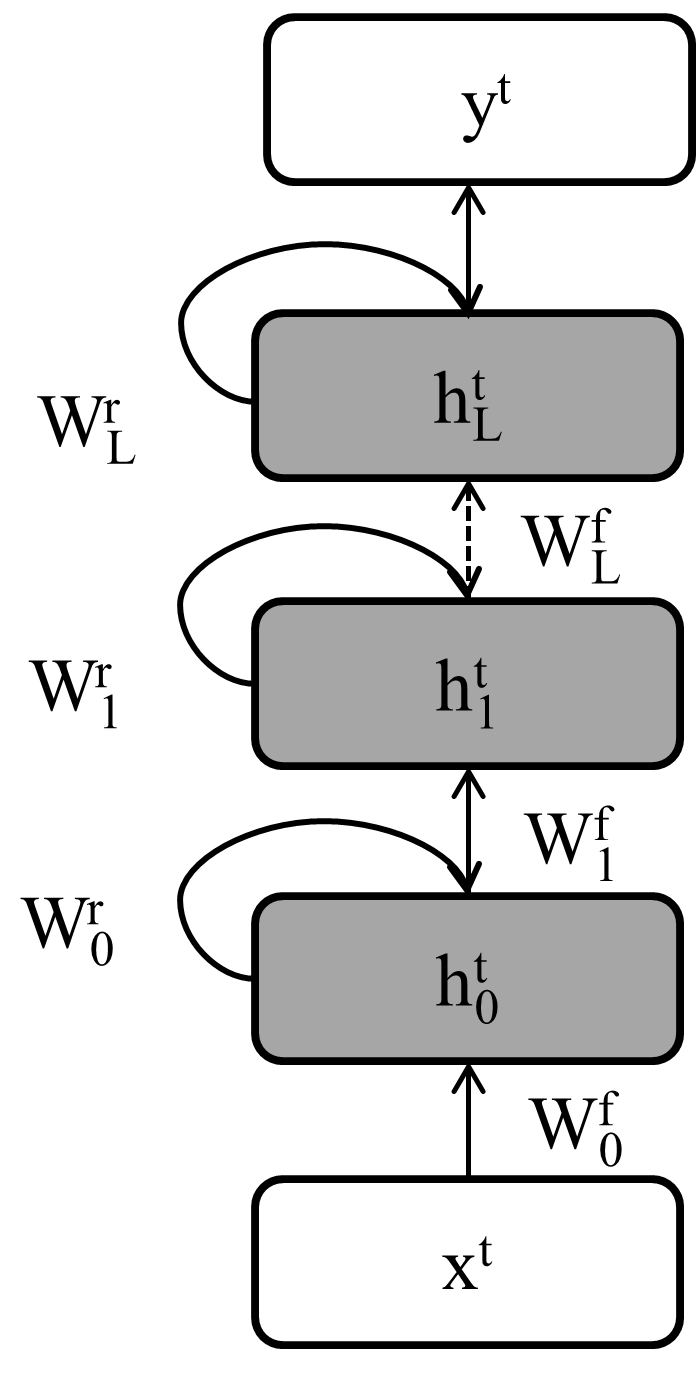}\label{RNN_fig}}
\hspace{30px}
\subfloat[ConvNet]{\includegraphics[width=0.65\textwidth,height=5cm]{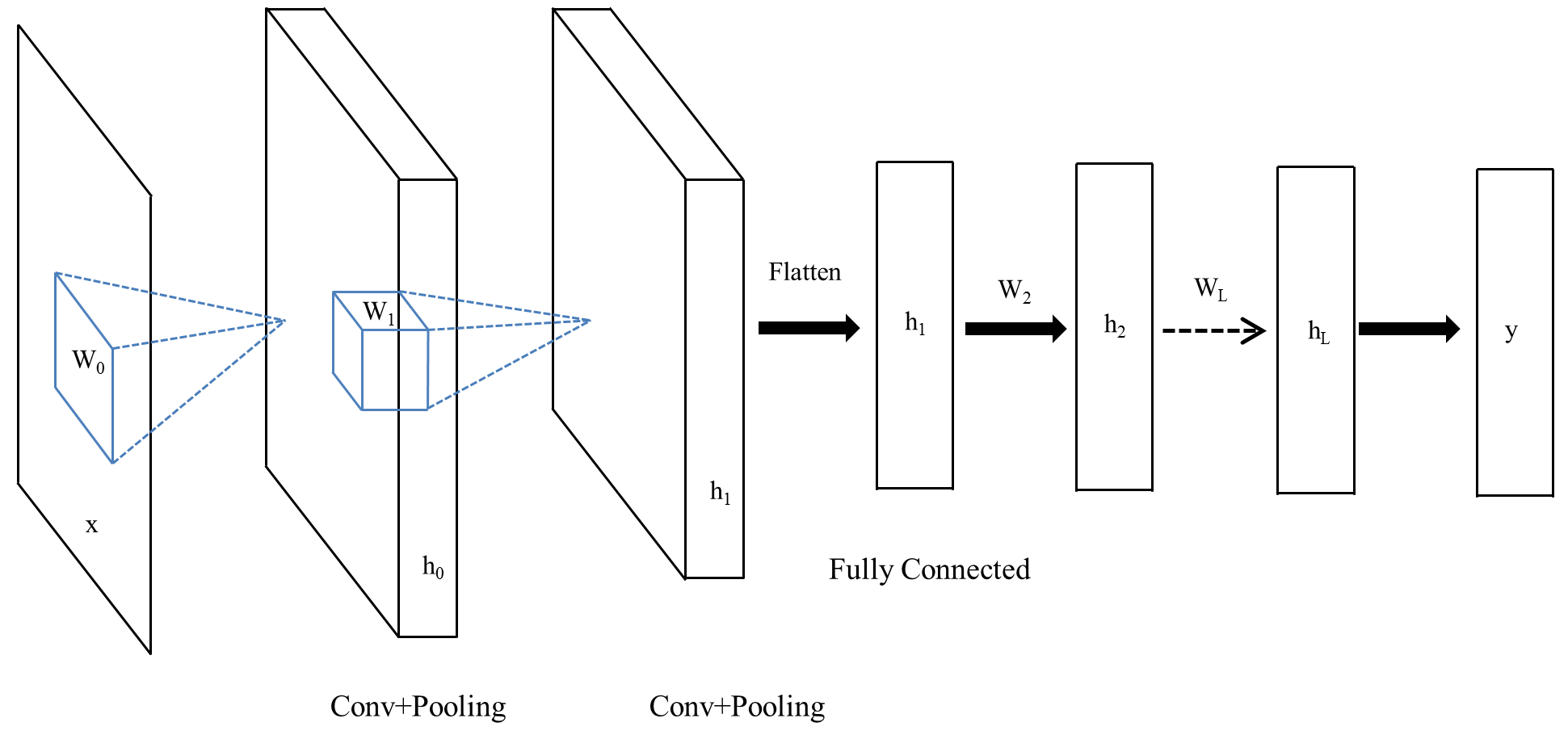}\label{CNN_fig}}
\caption{Neural network architectures for acoustic modelling.}
\label{architecture}
\end{figure*}

\section{Background}
\label{section:background}
In this section we describe the neural network models used for the acoustic and language modelling. Although neural networks are an old concept, they have recently been applied to a wide range of machine learning problems with great success \cite{lecun2015deep}. One of the primary reasons for their recent success has been the availability of large datasets and large-scale computing infrastructure \cite{dean2012large}, which makes it feasible to train networks with millions of parameters. The parameters of any neural network architecture are typically estimated with numerical optimisation techniques. Once a suitable cost function has been defined, the derivatives of the cost with respect to the model parameters are found using the backpropagation algorithm \cite{rumelhart1988learning} and parameters are updated using stochastic gradient descent (SGD) \cite{lecun2012efficient}. SGD has the useful property that the model parameters are iteratively updated using small batches of data. 
This allows the training algorithm to scale to very large datasets. The layered, hierarchical structure of neural nets makes end-to-end training possible, which implies that the network can be trained to predict outputs from low-level inputs without extracting features. This is in contrast to many other machine learning models whose performance is dependent on the features extracted from the data. Their ability to jointly learn feature transformations and classifiers makes neural networks particularly well suited to problems in MIR \cite{humphrey2013feature}.

\subsection{Acoustic Models}

\subsubsection{\textbf{Deep Neural Networks}}

DNNs are powerful machine learning models that can be used for classification and regression tasks. DNNs are characterised by having one or more layers of non-linear transformations. Formally, one layer of a DNN performs the following transformation:

\begin{equation}
\label{DNN}
h_{l+1} = f(W_l h_l + b_l). 
\end{equation}

In Equation \ref{DNN}, $W_l,b_l$ are the weight matrix and bias for layer $l$, $0 \leq l \leq L$ and $f$ is some non-linear function that is applied element-wise. For the first layer, $h_0 = x$, where $x$ is the input. In all our experiments, we fix $f$ to be the sigmoid function ($f(x) = \frac{1}{1+e^{-x}}$). The output of the final layer $h_L$ is transformed according to the given problem to yield a posterior probability distribution over the output variables $P(y|x,\theta)$. The parameters $\theta = \left\{ W_l,b_l\right\}^{L}_{0}$, are numerically estimated with the backpropagation algorithm and SGD. Figure \ref{DNN_fig} shows a graphical representation of the DNN architecture, the dashed arrows represent intermediate hidden layers. For acoustic modelling, the input to the DNN is a frame of features, for example a magnitude spectrogram or the constant Q transform (CQT) and the DNN is trained to predict the probability of pitches present in the frame $p(y_t|x_t)$ at some time $t$. 
 

\subsubsection{\textbf{Recurrent Neural Networks}}

DNNs are good classifiers for stationary data, like images. However, they are not designed to account for sequential data. RNNs are natural extensions of DNNs, designed to handle sequential or temporal data. This makes them more suited for AMT tasks, since consecutive frames of audio exhibit both short-term and long-term temporal patterns \cite{eck2002finding}.  RNNs are characterised by recursive connections between the hidden layer activations at some time $t$ and the hidden layer activations at $t-1$, as shown in Figure \ref{RNN_fig}. Formally, the hidden layer of an RNN at time $t$ performs the following computation:

\begin{equation}
\label{RNN}
h_{l+1}^t = f(W_l^f h_l^t + W_{l}^r h_l^{t-1} + b_l). 
\end{equation}

In Equation \ref{RNN}, $W_{l}^f$ is the weight matrix from the input to the hidden units, $W_{l}^r$ is the weight matrix for the recurrent connection and $b_l$ are the biases for layer $l$. From Equation \ref{RNN}, we can see that the recursive update of the hidden state at time $t$, implies that $h_t$ is implicitly a function of all the inputs till time $t$, $x_0^t$. Similar to DNNs, RNNs are made up of one or more layers of hidden units. The outputs of the final layer are transformed with a suitable function to yield the desired distribution over the ouputs. The RNN parameters $\theta = \left\{W_l^f,W_l^r,b_l\right\}^{L}_{0}$ are calculated using the back propagation through time algorithm (BPTT) \cite{werbos1990backpropagation} and SGD. For acoustic modelling, the RNN acts on a sequence of input features to yield a probability distribution over the outputs $P(y_t|x_0^t)$, where $x_0^t = \{ x_0,x_1,\ldots,x_t \}$. 

\subsubsection{\textbf{Convolutional Networks}}

ConvNets are neural nets with a unique structure. Convolutional layers are specifically designed to preserve the spatial structure of the inputs. In a convolutional layer, a set of weights act on a local region of the input. These weights are then repeatedly applied to the entire input to produce a \emph{feature map}. Convolutional layers are characterised by the sharing of weights across the entire input. As shown in Figure \ref{CNN_fig}, ConvNets are comprised of alternating convolutional and pooling layers, followed by one or more fully connected layers (same as DNNs). Formally, the repeated application of the shared weights to the input signal constitutes a convolution operation:
\begin{equation}
\label{CNN}
h_{j,k} = f( \sum_r W_{r,j} x_{r+k-1} + b_j). 
\end{equation}
The input $x$ is a vector of inputs from different channels, for example RGB channels for images. Formally, $x = \left\{x_0,x_1,\ldots\right\}$, where each input $x_i$ represents an input channel. Each input band $x_i$ has an associated weight matrix. All the weights of a convolutional layer are collectively represented as a four dimensional tensor. Given an $m \times n$ region from a feature map $h$, the max pooling function returns the maximum activation in the region. At any time $t$, the input to the ConvNet is a window of $2k+1$ feature frames $x_{t-k}^{t+k}$. The outputs of the final layer yield the posterior distribution distribution $P(y_t|x_{t-k}^{t+k})$. 

There are several motivations for using ConvNets for acoustic modelling. There are many experiments in MIR that suggest that rather than classifying a single frame of input, better prediction accuracies can be achieved by incorporating information over several frames of inputs \cite{sigtiachords,boulanger2013audio,bergstra2006aggregate}. Typically, this is achieved either by applying a context window around the input frame or by aggregating information over time by calculating statistical moments over a window of frames. Applying a context window around a frame of low level spectral features, like the short time fourier transform (STFT) would lead to a very high dimensional input, which is impractical. Secondly, taking mean, standard deviation or other statistical moments makes very simplistic assumptions about the distribution of data over time in neighbouring frames. 
ConvNets, due to their architecture \cite{lecun2015deep}, can be directly applied to several frames of inputs to learn features along both, the time and the frequency axes. 
Additionally, when using an input representation like the CQT, ConvNets can learn pitch-invariant features, since inter-harmonic spacings in music signals are constant across log-frequency. Finally, the weight sharing and pooling architecture leads to a reduction in the number of ConvNet parameters, compared to a fully connected DNN. This is a useful property given that very large quantities of labelled data are difficult to obtain for most MIR problems, including AMT. 

\subsection{Music Language Models}

Given a sequence $y = y_0^t$, we use the MLM to define a prior probability distribution $P(y)$. $y_t$ is a high-dimensional binary vector that represents the notes being played at $t$ (one time-step of a piano-roll representation). The high dimensional nature of the output space makes modelling $y_t$ a challenging problem. Most post-processing algorithms make the simplifying assumption that all the pitches are independent and model their temporal evolution with independent models \cite{poliner2007discriminative}. However, for polyphonic music, the pitches that are active concurrently are highly correlated (harmonies, chords). In this section, we describe the RNN music language models first introduced in \cite{boulanger2012modeling}. 

\subsubsection{\textbf{Generative RNN}}
\label{Gen_RNN_text}
The RNNs defined in the earlier sections were used to map a sequence of inputs $x$ to a sequence of outputs $y$. At each time-step $t$, the RNN outputs the conditional distribution $P(y_{t}|x_{0}^t)$. However RNNs can be used to define a distribution over some sequence $y$ by connecting the outputs of the RNN at $t-1$ to the inputs of the RNN at $t$, resulting in a distribution of the form: 

\begin{equation}
\label{genRNN}
P(y) = P(y_{0})\prod_{t > 0} P(y_t|y_0^{t-1})
\end{equation}

Although an RNN predicts $y_t$ conditioned on the high dimensional inputs $y_0^{t-1}$, the individual pitch outputs $y_t(i)$ are independent, where $i$ is the pitch index (Section \ref{preprocessing}). As mentioned earlier, this is not true for polyphonic music. Boulanger-Lewandowski et. al. \cite{boulanger2012modeling} demonstrate that rather than predicting independent distributions, the parameters of a more complicated parametric output distribution can be conditioned on the RNN hidden state. In our experiments, we use the RNN to output the biases of a neural autoregressive distribution estimator (NADE) \cite{boulanger2012modeling}. 

\subsubsection{\textbf{Neural Autogressive Distribution Estimator}}

The NADE is a distribution estimator for high dimensional binary data \cite{larochelle2011neural}. The NADE was initially proposed as a tractable alternative to the restricted Boltzmann machine (RBM). The NADE estimates the joint distribution over high dimensional binary variables as follows:
$$P(x) = \prod_i P(x_i|x_{0}^{i-1}).$$ 
The NADE is similar to a fully visible sigmoid belief network \cite{neal1992connectionist}, since the conditional probability of $x_i$ is a non-linear function of $x_0^t$. The NADE computes the conditional distributions according to:

\begin{equation}
h_i = \sigma(W_{:,<i}x_0^{i-1}+b_h)
\end{equation}

\begin{equation}
P(x_i|x_0^{i-1}) = \sigma(V_{i}h_{i} + b_{v}^{i})
\end{equation}
where $W,V$ are weight matrices, $W_{:,<i}$ is a submatrix of $W$ that denotes the first $i-1$ columns and $b_h,b_v$ are the hidden and visible biases, respectively. The gradients of the likelihood function $P(x)$ with respect to the model parameters $\theta = \left\{ W,V,b_h,b_v \right\}$ can be found exactly, which is not possible with RBMs \cite{larochelle2011neural}. This property allows the NADE to be readily combined with other models and the models can be jointly trained with gradient based optimisers. 

\subsubsection{\textbf{RNN-NADE}}

In order to learn high dimensional, temporal distributions for the MLM, we combine the NADE and an RNN, as proposed in \cite{boulanger2012modeling}. The resulting model yields a sequence of NADEs conditioned on an RNN, that describe a distribution over sequences of polyphonic music. The joint model is obtained by letting the parameters of the NADE at each time step be a function of the RNN hidden state $\theta_{NADE}^t = f(h_t)$. $h_t$ is the hidden state of final layer of the RNN (Equation \ref{RNN}) at time $t$.
In order to limit the number of free parameters in the model, we only allow the NADE biases to be functions of the RNN hidden state, while the remaining parameters ($W,V$) are held constant over time. We compute the NADE biases as a linear transformation of the RNN hidden state plus an added bias term \cite{boulanger2012modeling}:

\begin{equation}
b_v^t = b_v + W_1 h_t
\end{equation}

\begin{equation}
b_h^t = b_h + W_2 h_t
\end{equation}

$W_1$ and $W_2$ are weight matrices from the RNN hidden state to the visible and hidden biases, respectively. The gradients with respect to all the model parameters can be easily computed using the chain rule and the joint model is trained using the BPTT algorithm \cite{boulanger2012modeling}. 

\section{Proposed Model}
\label{section:methods}

In this section we review the proposed neural network model for polyphonic AMT. As mentioned earlier, the model comprises an acoustic model and a music language model. In addition to the acoustic models in \cite{sigtia2014hybrid}, we propose the use of ConvNets for identifying pitches present in the input audio signal and compare their performance to various other acoustic models (Section \ref{section:results}). The acoustic and language models are combined under a single training objective using a hybrid RNN architecture, yielding an end-to-end model for AMT with unconstrained polyphony. We first describe the hybrid RNN model, followed by a description of the proposed inference algorithm. 


\subsection{Hybrid RNN}

The hybrid RNN is a graphical model that combines the predictions of any \emph{arbitrary} frame level acoustic model, with an RNN-based language model. Let $x = x_{0}^T$ be a sequence of inputs and let $y = y_{0}^T$ be the corresponding transcriptions. The joint probability of $y,x$ can be factorised as follows:

\begin{align}\label{factorisation}
 P(y,x) &= P(y_{0} \ldots y_{T} , x_{0} \ldots x_{T}) \\ \nonumber
  &= P(y_0)P(x_{0}|y_0) \prod_{t=1}^{T} P(y_t|y_{0}^{t-1}) P(x_t|y_t). \\ \nonumber
\end{align}
The factorisation in Equation \ref{factorisation} makes the following independence assumptions:

\begin{equation}
P(y_t|y_{0}^{t-1},x_{0}^{t-1}) = P(y_t|y_{0}^{t-1})
\end{equation}

\begin{equation}
\label{independence}
P(x_t|y_{0}^{t},x_{0}^{t-1}) = P(x_t|y_t)
\end{equation}

\begin{figure}[!t]
\centering
\includegraphics[height=4cm]{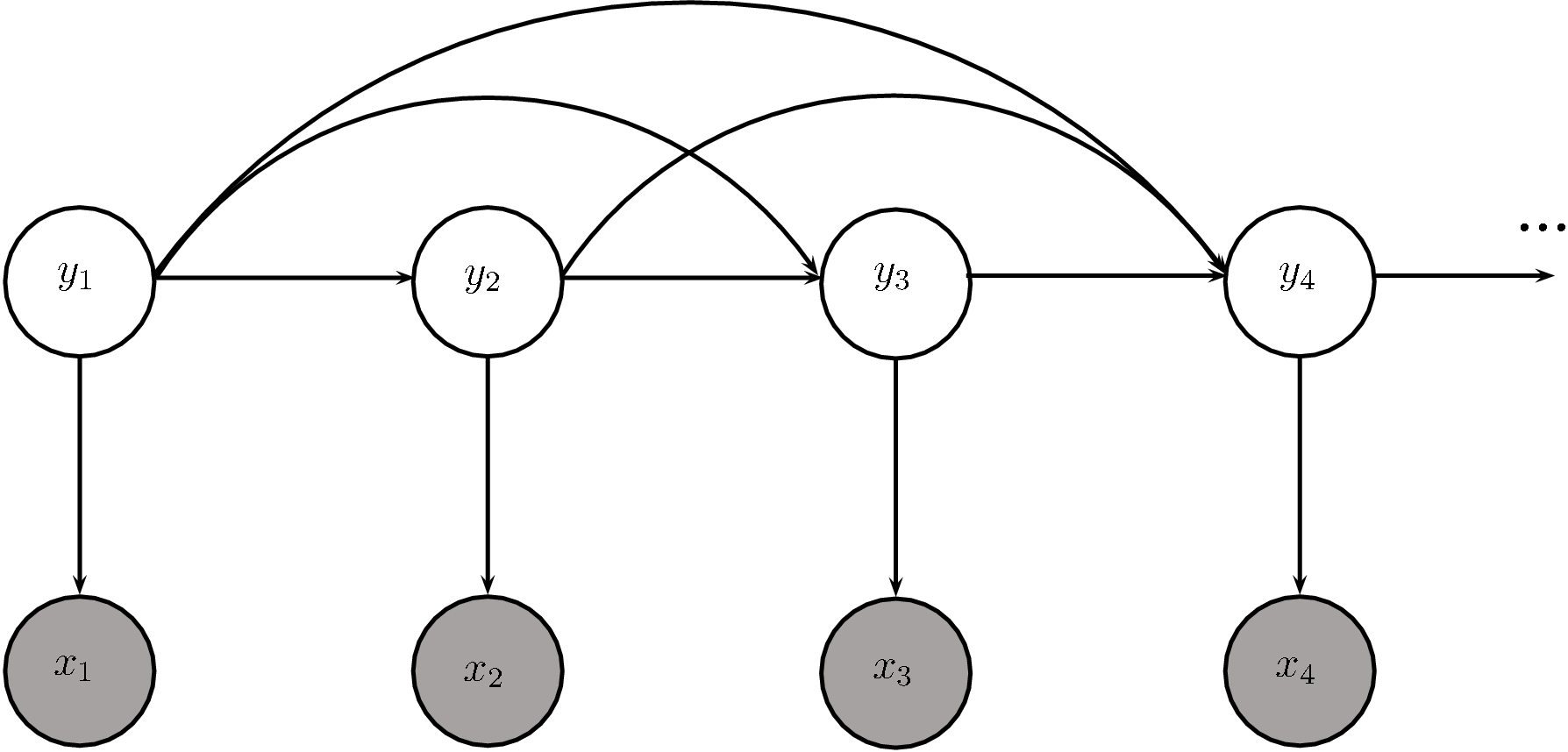}
\caption{Graphical Model of the Hybrid Architecture}
\label{fig_hybrid}
\end{figure}

These independence assumptions are similar to the assumptions made in HMMs \cite{rabiner1989tutorial}. Figure \ref{fig_hybrid} is a graphical representation of the hybrid model. In equation \ref{factorisation}, $P(x_t|y_t)$ is the \emph{emission} probability of an input, given output $y_t$. Using Bayes's rule, the conditional distribution can be written as follows:

\begin{equation}
\label{joint_prob}
P(y|x) \propto P(y_0|x_0)\prod_{t=1}^{T} P(y_t|y_{0}^{t-1}) P(y_t|x_t),
\end{equation}
where the marginals $P(y_t)$ and priors $P(y_0)$, $P(x_0)$ are assumed to be fixed w.r.t. the model parameters.  

With this reformulation of the joint distribution, we observe that the conditional distribution $P(y|x)$ is directly proportional to the product of two distributions. The prior distribution $P(y_t|y_{0}^{t-1})$ is obtained using a generative RNN (Section \ref{Gen_RNN_text}) and the posterior distribution over note-combinations $P(y_t|x_t)$ can be modelled using any frame based classifier. The hybrid RNN graphical model is similar to an HMM, where the state transition probabilities for the HMM $P(y_t|y_{t-1})$ have been generalised to include connections from all previous outputs, resulting in the $P(y_t|y_{0}^{t-1})$ terms in Equation \ref{joint_prob}.


For the problem of automatic music transcription, the input time-frequency representation forms the input sequence $x$, while the output piano-roll sequence $y$ denotes the transcriptions. The priors $P(y_t|y_{0}^{t-1})$ are obtained from the RNN-NADE MLM, while the posterior distributions $P(y_t|x_t)$ are obtained from the acoustic models. The models can then be trained by finding the derivatives of the acoustic and language model objectives with respect to the model parameters and training using gradient descent. The independent training of the acoustic and language models is a useful property since datasets available for music transcription are considerably smaller in size as compared to datasets in computer vision and speech. However large corpora of MIDI music are relatively easy to find on the internet. Therefore in theory, the MLMs can be trained on large corpora of MIDI music, analogous to language model training in speech. 




\subsection{Inference}

At test time, we would like to find the mode of the conditional output distribution:
\begin{equation}
\label{inference}
y^* = \argmax_y P(y|x)
\end{equation}
From Equation \ref{joint_prob}, we observe that the priors $P(y_t|y_{0}^{t-1})$, tie the predictions of the acoustic model $P(y_t|x_t)$ to all the predictions made till time $t$. This prior term encourages coherence between predictions over time and allows musicological structure learnt by the language models to influence successive predictions. However, this more general structure leads to a more complex inference (or decoding) procedure at test time. This is due to the fact that at time $t$, the history $y_{0}^{t-1}$ has not been optimally determined. Therefore, the optimum choice of $y_t$ depends on \emph{all} the past model predictions. Proceeding greedily in a chronological manner by selecting $y_t$ that optimises $P(y_t|x_t)$ does not necessarily yield good solutions. 
We are interested in solutions that globally optimise $p(y|x)$. But exhaustively searching for the best sequence is intractable since the number of possible configurations of $y_t$ is exponential in the number of output pitches ($2^n$ for $n$ pitches).

Beam search is a graph search algorithm that is commonly used to decode the conditional outputs of an RNN \cite{graves2012sequence,boulanger2013high,sigtiachords}. Beam search scales to arbitrarily long sequences and the computational cost versus accuracy trade-off can be controlled via the width of the beam. The inference algorithm is comprised of the following steps: at any time $t$, the algorithm maintains at most $w$ partial solutions, where $w$ is the beam width or the beam capacity. The solutions in the beam at $t$ correspond to sub-sequences of length $t$. Next, all possible descendants of the $w$ partial solutions in the beam are enumerated and then sorted in decreasing order of log-likelihood. From these candidate solutions, the top $w$ solutions are retained as beam entries for further search. Beam search can be readily applied to problems where the number of candidate solutions at each step is limited, like speech recognition \cite{boulanger2014phone} and audio chord estimation \cite{sigtiachords}.
 However, using beam search for decoding sequences with a large output space is prohibitively inefficient.


When the space of candidate solutions is large, the algorithm can be constrained to consider only $K$ new candidates for each partial solution in the beam, where $K$ is known as the \emph{branching factor}. The procedure for selecting the $K$ candidates can be designed according to the given problem. For the hybrid architecture, from Equation \ref{joint_prob} we note:

\begin{equation}
\label{beam_step}
P(y_0^t|x_0^t) \propto P(y_0^{t-1}|x_0^{t-1}) P(y_t|y_{0}^{t-1})P(y_t|x_t)
\end{equation}

At time $t$, the partial solutions in the beam correspond to configurations of $y_0^{t-1}$. Therefore given $P(y_0^{t-1}|x_0^{t-1})$, the $K$ configurations that maximise $P(y_t|y_{0}^{t-1})P(y_t|x_t)$ would be a suitable choice of candidates for $y_t$. However for many families of distributions, it might not be possible to enumerate $y_t$ in decreasing order of likelihood. In \cite{boulanger2013high}, the authors propose forming a pool of $K$ candidates by drawing random samples from the conditional output distributions. However, random sampling can be inefficient and obtaining independent samples can be very expensive for many types of distributions. As an alternative, we propose to sample solutions from the posterior distribution of the acoustic model $P(y_t|x_t)$ \cite{sigtia2014hybrid}. 
There are $2$ main motivations for doing this. Firstly, the outputs of the acoustic model are independent class probabilities. Therefore, it is easy to enumerate samples in decreasing order of log-likelihood \cite{boulanger2013high}. Secondly, we avoid the accumulation of errors in the RNN predictions over time \cite{bengio2015scheduled}. The RNN models are trained to predict $y_t$, given the \emph{true} outputs $y_0^{t-1}$. However at test time, outputs sampled from the RNN are fed back as inputs at the next time step. This discrepancy between the training and test objectives can cause prediction errors to accumulate over time.

\begin{algorithm}
\caption{High Dimensional Beam Search}
\begin{algorithmic}
\State{Find the most likely sequence $y$ given $x$ with a beam width $w$ and branching factor $K$.}
\State{$beam \gets$ new beam object}
\State{$beam$.insert$(0,\left\{\right\})$}
\For{$t$ = $1$ to $T$}
  \State{$new\_beam \gets$ new beam object}
  \For{$l,s,m_{a},m_{l}$ \textbf{in} $beam$}
        \For{$k=1$ to $K$ }
          \State{$y' = m_{a}$.next\_most\_probable()}
          \State{$l' = \log P_{l}(y'|s)P_{a}(y'|x_t)-\log P(y')$}
          \State{$m_{l}' \gets m_{l}$ with $y_t := y'$}
          \State{$m_{a}' \gets m_{a}$ with $x := x_{t+1}$}
          \State{$new\_beam$.insert($l+l',\left\{ s,y' \right\}$,$m_a$,$m_l$)}
    \EndFor
  \EndFor
\vspace{-0.15cm}
  \State{$beam \gets new\_beam$}
\vspace{0.1cm}
\EndFor
\Return{$beam.$pop()}
\end{algorithmic}
\end{algorithm}

Although generating candidates from the acoustic model yields good results, it requires the use of large beam widths. This makes the inference procedure computationally slow and unsuitable for real-time applications \cite{sigtia2014hybrid}. In this study, we propose using the \emph{hashed beam search} algorithm proposed in \cite{sigtiachords}. Beam search is fundamentally limited when decoding long temporal sequences. This is due to the fact that solutions that differ at only a few time-steps, can saturate the beam. This causes the algorithm to search a very limited space of possible solutions. 
This issue can be solved by efficient pruning. The hashed beam search algorithm improves efficiency by pruning solutions that are \emph{similar} to solutions with a higher likelihood. The metric that determines the similarity of sequences can be chosen in a problem dependent manner and is encoded in the form of a locality sensitive hash function \cite{sigtiachords}. 
In Algorithm $1$, we outline the beam search algorithm algorithm used for our experiments, while Algorithm $2$ describes the hash table beam object. In Algorithms $1$ and $2$, $s$ is a sequence $y_0^t$, $l$ is log-likelihood of $s$, $m_a,m_l$ are acoustic and language model objects and $f_h$ is the hash function.


There are two key differences between Algorithm $1$ and the algorithm in \cite{sigtia2014hybrid}. First, the priority queue that stores the beam is replaced by a hash table beam object (see Algorithm $2$). Secondly, for each entry in the beam we evaluate $K$ candidate solutions. This is in contrast to the algorithm in \cite{sigtia2014hybrid}, where once the beam is full, only $w$ candidate solutions are evaluated per iteration. It might appear that the hashed beam search algorithm might be more expensive, since it evaluates $w*K$ candidates instead of $w$ candidates. However, by efficiently pruning similar solutions, the algorithm yields better results for much smaller values of $w$, resulting in a significant increase in efficiency (Section \ref{section:results}, Figure \ref{beam_search}). 

\begin{algorithm}
\caption{Description of beam objects given $w,f_h,k$}
\begin{algorithmic}
\State{\textbf{Initialise beam object}}
  \State{beam.hashQ = defaultdict of priority queues$^{*}$}
  \State{beam.queue = indexed priority queue of length $w$$^{**}$}
\State{\textbf{Insert $l,s$ into beam}}
  \State{key$=f_h(s)$}
  \State{queue = beam.queue}
  \State{hashQ = beam.hashQ[key]}
  \State{fits\_in\_queue = \textbf{not} queue.full() \textbf{or} $l\geq$queue.min()}
  \State{fits\_in\_hashQ = \textbf{not} hashQ.full() \textbf{or} $l\geq$hashQ.min()}
\If{fits\_in\_queue \textbf{and} fits\_in\_hashQ}
    \State{hashQ.insert($l,s$)}
    \If{hashQ.overfull()}
      \State{item = hashQ.del\_min()}
      \State{queue.remove(item)}
    \EndIf
    \State{queue.insert($l,s$)}
    \If{queue.overfull()}
      \State{item = queue.del\_min()}
      \State{beam.hashQ[$f_h$(item.$s$)].remove(item)}
    \EndIf
\EndIf
\State{$^{*}$ A priority queue of length $k$ maintains the top $k$ entries at all times.} 
\State{$^{**}$ An \emph{indexed} priority queue allows efficient random access and deletion.}
\end{algorithmic}
\end{algorithm}

Algorithm $2$ describes the hash table beam object. The hashed beam search algorithm offers several advantages compared to the standard beam search algorithm. The notion of similarity of solutions can be encoded in the form of hash functions. For music transcription, we choose the similarity function to be the last $n$ frames in a sequence $s$. $n=1$ corresponds to a dynamic programming like decoding (similar to HMMs) where all sequences with the same final state $y_t$ are considered to be equivalent, and the sequence with the highest log-likelihood is retained. $n = $ len(sequence) corresponds to regular beam search. Additionally, the hash beam search algorithm can maintain $\geq 1$ solution per hash key through a process called chaining \cite{cormen2001introduction}. 


\section{Evaluation}
\label{section:evaluation}
In this section we describe how the performance of the proposed model is evaluated for a polyphonic transcription task. 
\subsection{Dataset}
\label{dataset}
We evaluate the proposed model on the MAPS dataset \cite{emiya2010multipitch}. The dataset consists of audio and corresponding annotations for isolated sounds, chords and complete pieces of piano music. For our experiments, we use only the full musical pieces for training and testing the neural network acoustic models and MLMs. The dataset consists of $270$ pieces of classical music and MIDI annotations. There are $9$ categories of recordings corresponding to different piano types and recording conditions, with $30$ recordings per category. $7$ categories of audio are produced by software piano synthesisers, while $2$ sets of recordings are obtained from a Yamaha Disklavier upright piano. Therefore the dataset consists of $210$ synthesised recordings and $60$ real recordings.  


We perform 2 sets of investigations in this paper. The first set of experiments investigate the effect of the RNN MLMs on the predictions of the acoustic models. For this task, we divide the entire dataset set into 4 disjoint train/test splits, as to ensure that the folds are music piece-independent. Specifically, for some of the musical pieces in the dataset, audio for each piece is rendered using more than one piano. Therefore while creating the splits, we ensure that the training and test data do not contain any overlapping pieces\footnote{Details available at: \url{http://www.eecs.qmul.ac.uk/~sss31/TASLP/info.html}}. For each split, we select $80\%$ of the data for training ($216$ musical pieces) and the remaining for testing ($54$ pieces). From each training split, we hold out $26$ tracks as a validation set for selecting the hyper-parameters for the training algorithm (Section \ref{section:training}). All the reported results are mean values of the evaluation metrics over the $4$ splits. From now on, 
this evaluation configuration will be named as \emph{Configuration 1}.

Although the above experimental setup is useful for investigating the effectiveness of the RNN MLMs, the training set contains examples from piano models which are used for testing. This is usually not true in practice, where the instrument models/sources at test time are unknown and usually do not coincide with the instruments used for training. A majority of experiments with the MAPS dataset train and test model on disjoint instrument types \cite{benetos2012shift,berg2014unsupervised,o2014polyphonic}. We thus perform a second set of experiments to compare performance of the different neural network acoustic models in a more realistic setting. We train the acoustic models using the 210 tracks created using synthesized pianos (180 tracks for training and 30 tracks for validation) and we test the acoustic models on the 60 audio recordings obtained from Yamaha Disklavier piano recordings (models `ENSTDkAm' and `ENSTDkCl' in the MAPS database). In this experiment, we do not apply the language models since the 
train and test sets contain overlapping musical pieces. In addition to the neural network acoustic models, we include comparisons with two state-of-the-art unsupervised acoustic models \cite{benetos2012shift, vincent2010adaptive} for both experiments. This instrument source-independent evaluation configuration will be named from now on as \emph{Configuration 2}.

\subsection{Metrics}

We use both frame and note based metrics to assess the performance of the proposed system \cite{bay2009evaluation}. Frame-based evaluations are made by comparing the transcribed binary output and the MIDI ground truth frame-by-frame. For note-based evaluation, the system returns a list of notes, along with the corresponding pitches, onset and offset time. We use the F-measure, precision, recall and accuracy for both frame and note based evaluation. Formally, the frame-based metrics are defined as:

$$ \mathcal{P} = \sum_{t=1}^{T} \frac{TP[t]}{TP[t]+FP[t]}$$

$$ \mathcal{R} = \sum_{t=1}^{T} \frac{TP[t]}{TP[t]+FN[t]}$$

$$ \mathcal{A} = \sum_{t=1}^{T} \frac{TP[t]}{TP[t]+FP[t]+FN[t]}$$

$$ \mathcal{F} = \frac{2*\mathcal{P}*\mathcal{R}}{\mathcal{P}+\mathcal{R}}$$

where TP[t] is the number of true positives for the event at $t$, FP is the number of false positives and FN is the number of false negatives. The summation over $T$ is carried out over the entire test data. Similarly, analogous note-based metrics can be defined \cite{bay2009evaluation}. A note event is assumed to be correct if its predicted pitch onset is within a $\pm 50$ $ms$ range of the ground truth onset. 

\subsection{Preprocessing}
\label{preprocessing}
We transform the input audio to a time-frequency representation which is then input to the acoustic models. In \cite{sigtia2014hybrid}, we used the magnitude short-time Fourier transform (STFT) as input to the acoustic models. However, here we experiment with the constant Q transform (CQT) as the input representation. There are two motivations for this. Firstly, the CQT is fundamentally better suited as a time-frequency representation for music signals, since the frequency axis is linear in pitch \cite{brown1991calculation}. Another advantage of using the CQT is that the resulting representation is much lower dimensional than the STFT. Having a lower dimensional representation is useful when using neural network acoustic models as it reduces the number of parameters in the model. 

We downsample the audio to $16$ kHz from $44.1$ kHz. We then compute CQTs over $7$ octaves with $36$ bins per octave and a hop size of $512$ samples, resulting in a $252$ dimensional input vector of real values, with a frame rate of $31.25$ frames per second. Additionally, we compute the mean and standard deviation of each dimension over the training set and transform the data by subtracting the mean and diving by the standard deviation. These pre-processed vectors are used as inputs to the acoustic model. For the language model training, we sample the MIDI ground truth transcriptions of the training data at the same rate as the audio ($32$ ms). We obtain sequences of $88$ dimensional binary vectors for training the RNN-NADE language models. The $88$ outputs correspond to notes A0-C8 on a piano. 

The test audio is sampled at a frame rate of $100$ Hz yielding $100*30 = 3000$ frames per test file. For $54$ test files over $4$ splits, we obtain a total of $648,000$ frames at test time\footnote{It should be noted that carrying out statistical significance tests on a track level is an over-simplification in the context of multi-pitch detection, as argued in \cite{BenetosThesis}.}. 

\subsection{Network Training}
\label{section:training}
In this section we describe the details of the training procedure for the various acoustic model architectures and the RNN-NADE language model. All the acoustic models have $88$ units in the output layer, corresponding to the $88$ output pitches. The outputs of the final layer are transformed by a sigmoid function and yield independent pitch probabilities $P(y_t(i) = 1 | x)$. All the models are trained by maximising the log-likelihood over all the examples in the training set. 

\subsubsection{\textbf{DNN Acoustic Models}}

For DNN training, we constrain all the hidden layers of the model to have the same number of units to simplify searching for good model architectures. We perform a grid search over the following parameters: number of layers $L \in \left\{ 1,2,3,4 \right\}$, number of hidden units $H \in \left\{ 25,50,100,125,150,200,250 \right\}$, hidden unit activations $act \in \left\{ ReLU, sigmoid \right\}$ where ReLU is the rectified linear unit activation function \cite{glorot2011deep}. We found Dropout \cite{srivastava2014dropout} to be essential for improving generalisation performance. A Dropout rate of $0.3$ was used for the input layer and all the hidden layers of the network. 
Rather than using learning rate and momentum update schedules, we use ADADELTA \cite{zeiler2012adadelta} to adapt the learning over iterations. In addition to Dropout, we use early stopping to minimise overfitting. Training was stopped if the cost over the validation set did not decrease for $20$ epochs. We used mini batches of size $100$ for the SGD updates.

\subsubsection{\textbf{RNN Acoustic Models}}

For RNN training, we constrain all the hidden layers to have the same number of units. We perform a grid search over the following parameters: $L \in \left\{ 1,2,3 \right\} $, $H \in \left\{ 25,50,100,150,200,250 \right\}$. We fix the hidden activations of the recurrent layers to be the hyperbolic tangent function. We found that ADADELTA was not particularly well suited for training RNNs. We use an initial learning rate of $0.001$ and linearly decrease it to $0$ over $1000$ iterations. We use a constant momentum rate of $0.9$.
The training sequences are further divided into sub-sequences of length $100$. The SGD updates are made one sub-sequence at a time, without any mini batching. Similar to the DNNs, we use early stopping and stop training if validation cost does not decrease after $20$ iterations. In order to prevent gradient explosion in the early stages of training, we use gradient clipping \cite{bengio2013advances}. We clipped the gradients, when the norm of the gradient was greater than 5.  

\subsubsection{\textbf{ConvNet Acoustic Models}}

The input to the ConvNet is a context window of frames and the target is the central frame in the window \cite{sigtiachords}. The frames at the beginning and end of the audio are zero padded so that a context window can be applied to each frame. Although pooling can be performed along both axes, we only perform pooling over the frequency axis. We performed a grid search over the following parameters: window size $ w_s \in \left\{ 3,5,7,9 \right\}$ number of convolutional layers $L_c \in \left\{ 1,2,3,4 \right\}$, number of filters per layer $n_l \in \left\{ 10,25,50,75,100 \right\}$, number of fully connected layers $L_{fc} \in \left\{ 1,2,3 \right\}$, number of hidden units in fully connected layers $H \in \left\{ 200,500,1000 \right\}$. The convolution activation functions were fixed to be the hyperbolic tangent functions, while all the fully connected layer activations were set to the sigmoid function. The pooling size is fixed to be $P = (1,3)$ for all convolutional layers. Dropout with rate $0.5$ is 
applied to all convolutional layers. 
We tried a large permutation of window shapes for the convolutional layer and the following subset of window shapes yielded good results: $w \in \left\{ (3,3),(3,5),(5,5),(3,25),(5,25),(3,75),(5,75) \right\}$. We observed that classification performance deteriorated sharply for longer filters along the frequency axis. $0.5$ Dropout was applied to all the fully connected layers. The model parameters were trained with SGD and a batch size of $256$. An initial learning rate of $0.01$ was linearly decreased to $0$ over $1000$ iterations. A constant momentum rate $0.9$ was used for all the updates. We stopped training if the validation error did not decrease after $20$ iterations over the entire training set.  

\begin{table*}[htpb]
\begin{center}
\scalebox{1.2}{
  \begin{tabular}{|c| c  c | c  c | c c|}
    \hline
    Post Processing&\multicolumn{2}{c|}{Thresholding}&\multicolumn{2}{c|}{HMM}&\multicolumn{2}{c|}{Hybrid Architecture}\\ \hline
    Acoustic Model & Frame & Note & Frame & Note & Frame & Note \\ \hline
    Benetos \cite{benetos2012shift}  & $64.20$ & $65.22$ & $64.84$ & $66.05$ & $65.10$ & $66.48$ \\ \hline
    Vincent \cite{vincent2010adaptive}  & $58.95$ & $\mathbf{68.5}$ & $60.37$ & $\mathbf{68.87}$ & $59.78$ & $\mathbf{69.00}$ \\ \hline 
    DNN & $67.54$ & $60.02$ & $68.32$ & $62.26$ & $67.92$ & $63.18$ \\ \hline
    RNN & $68.38$ & $63.84$ & $68.09$ & $64.50$ & $69.25$ & $65.24$ \\ \hline
    ConvNet & $\mathbf{73.57}$ & $65.35$ & $\mathbf{73.75}$ & $66.20$ & $\mathbf{74.45}$ & $67.05$\\ \hline
  \end{tabular}}
\end{center}
\caption{F-measures for multiple pitch detection on the MAPS dataset, using evaluation configuration 1.}
\label{f-scores}
\end{table*}

\begin{table*}[htpb]
\begin{center}
\scalebox{1.2}{
  \begin{tabular}{|c| c  c | c  c | c  c |}
    \hline
    & \multicolumn{2}{|c|}{$\mathcal{P}$}&\multicolumn{2}{c|}{$\mathcal{R}$}&\multicolumn{2}{c|}{$\mathcal{A}$} \\ \hline
    Acoustic Model& Frame & Note & Frame & Note & Frame & Note \\ \hline
    Benetos \cite{benetos2012shift} & $59.54$ & $73.51$ & $69.51$ & $60.67$ & $48.47$ & $49.03$  \\ \hline
    Vincent \cite{vincent2010adaptive} & $52.71$ & $\mathbf{79.93}$ & $69.04$ & $60.69$ & $43.04$ & $\mathbf{52.92}$  \\ \hline 
    DNN & $65.66$ & $62.62$ & $70.34$ & $63.75$ & $51.76$ & $45.33$  \\ \hline
    RNN & $67.89$ & $64.64$ & $70.66$ & $65.85$ & $54.38$ & $48.18$  \\ \hline
    ConvNet & $\mathbf{72.45}$ & $67.75$ &$\mathbf{76.56}$ & $\mathbf{66.36}$ & $\mathbf{58.87}$ & $50.07$  \\ \hline
  \end{tabular}}
\end{center}
\caption{Precision, Recall and Accuracy for multiple pitch detection on the MAPS dataset using the hybrid architecture ($w=10,K=4,k=2,f_h(y_0^t)=y_t$), using evaluation configuration 1.}
\label{more-metrics}
\end{table*}

\begin{table*}[htpb]
\begin{center}
\scalebox{1.2}{
  \begin{tabular}{|c| c | c | c | c | c |}
    \hline
    Acoustic Model & Benetos \cite{benetos2012shift} & Vincent \cite{vincent2010adaptive} & DNN & RNN & ConvNet \\ \hline
    F-measure (Frame) & $59.31$ & $59.60$ & $59.91$ & $57.67$ & $\mathbf{64.14}$ \\ \hline
    F-measure (Note) & $54.29$ & $\mathbf{59.12}$ & $49.43$ & $49.20$ & $54.89$ \\ \hline
  \end{tabular}}
\end{center}
\caption{F-measures for acoustic models trained on synthesised pianos and tested on real recordings (evaluation configuration 2).}
\label{Disklavier}
\end{table*}

\subsubsection{\textbf{RNN-NADE Language Models}}

The RNN-NADE models were trained with SGD and with sequences of length $100$. We performed a grid search over the following parameters: number of recurrent units $H_{RNN} \in \left\{ 50,100,150,200,250,300 \right\}$ and number of hidden units for the NADE $H_{NADE} \in \left\{ 50,100,150,200,250,300 \right\}$. The model was trained with an initial learning rate of $0.001$ which was linearly reduced to $0$ over $1000$ iterations. A constant momentum rate of $0.9$ was applied throughout training. 

We selected the model architectures by performing a grid search over the parameter values described earlier in the section. The various models were evaluated on one train/test split and the best performing architecture was then used for all other experiments. 

\subsection{Comparative Approaches}
\label{section:comparison}
For comparative purposes, two state-of-the-art polyphonic music transcription methods were used for experiments \cite{benetos2012shift, vincent2010adaptive}. In both cases, the non-binary \emph{pitch activation} output of the aforementioned methods was extracted, for performing an in-depth comparison with the proposed neural network models. 
The multi-pitch detection method of \cite{vincent2010adaptive} is based on non-negative matrix factorization (NMF) and operates by decomposing an input time-frequency representation as a series of basis spectra (representing pitches) and component activations (indicating pitch activity across time). This method models each basis spectrum as a weighted sum of narrowband spectra representing a few adjacent harmonic partials, enforcing harmonicity and spectral smoothness. As input time-frequency representation, an Equivalent Rectangular Bandwidth (ERB) filterbank is used. Since the method relies on a dictionary of (hand-crafted) narrowband harmonic spectra, system parameters remain the same for the two evaluation configurations.

The multiple-instrument transcription method of \cite{benetos2012shift} is based on shift-invariant PLCA (a convolutive and probabilistic counterpart of NMF). In this model, the input time-frequency representation is decomposed into a series of basis spectra per pitch and instrument source which are shifted across log-frequency, thus supporting tuning changes and frequency modulations. Outputs include the pitch activation distribution and the instrument source contribution per pitch. Contrary to the parametric model of \cite{vincent2010adaptive}, the basis spectra are pre-extracted from isolated musical instrument sounds. 
As in the proposed method, the input time-frequency representation of \cite{benetos2012shift} is the CQT. For the investigations with MLMs (configuration 1), the PLCA models are trained on isolated sound examples from all 9 piano models from the MAPS database (in order for the experiments to be comparable with the proposed method). For the second set of experiments which investigate the generalisation capabilities of the models (configuration 2), the PLCA acoustic model is trained on isolated sounds from the sysnthesised pianos and tested on recordings created using the Yamaha Disklavier piano.


\subsection{Results}
\label{section:results}

In this section we present results from the experiments on the MAPS dataset. As mentioned before, all results are the mean values of various metrics computed over the $4$ different train/test splits. The acoustic models yield a sequence of probabilities for the individual pitches being active (posteriograms). The post-processing methods are used to transform the posteriograms to a binary piano-roll representation. The various performance metrics (both frame and note based) are then computed by comparing the outputs of the systems to the ground truth. 

\begin{table}[t!]
\begin{center}
\scalebox{0.9}{
\begin{tabular}{|c|c|}
  \hline
  Model & Architecture \\ \hline
  DNN & $L=3,H=125$ \\ \hline
  RNN & $L=2,H=200$ \\ \hline
  ConvNet & $w_s=7,L_c=2,L_{fc}=2,w_1 = (5,25),P_1 = (1,3)$ \\ 
  & $w_2=(3,5),P_2=(1,3),n_1=n_2=50,h_1=1000,h_2=200$ \\ \hline
  RNN-NADE & $H_{RNN}=200,H_{NADE}=150$\\ \hline
\end{tabular}}
\end{center}
\caption{Model configurations for the best performing architectures. }
\label{model_architectures}
\end{table}

We consider $3$ kinds of post-processing methods. The simplest post-processing method is to apply a threshold to the output pitch probabilities obtained from the acoustic model. We select the threshold that maximises the F-measure over the entire training set and use this threshold for testing. Pitches with probabilities greater than the threshold are set to 1, while the remaining pitches are set to 0. The second post-processing method considered uses individual pitch HMMs for post-processing similar to \cite{poliner2007discriminative}. The HMM parameters (transition probabilities, pitch marginals) are obtained by counting the frequency of each event over the MIDI ground truth data. The binary pitch outputs are obtained using Viterbi decoding \cite{rabiner1989tutorial}, where the scaled likelihoods are used as emission probabilities. Finally, we combine the acoustic model predictions with the RNN-NADE MLMs and obtain binary transcriptions using beam search.

\begin{figure}[t!]
\includegraphics[width=0.5\textwidth]{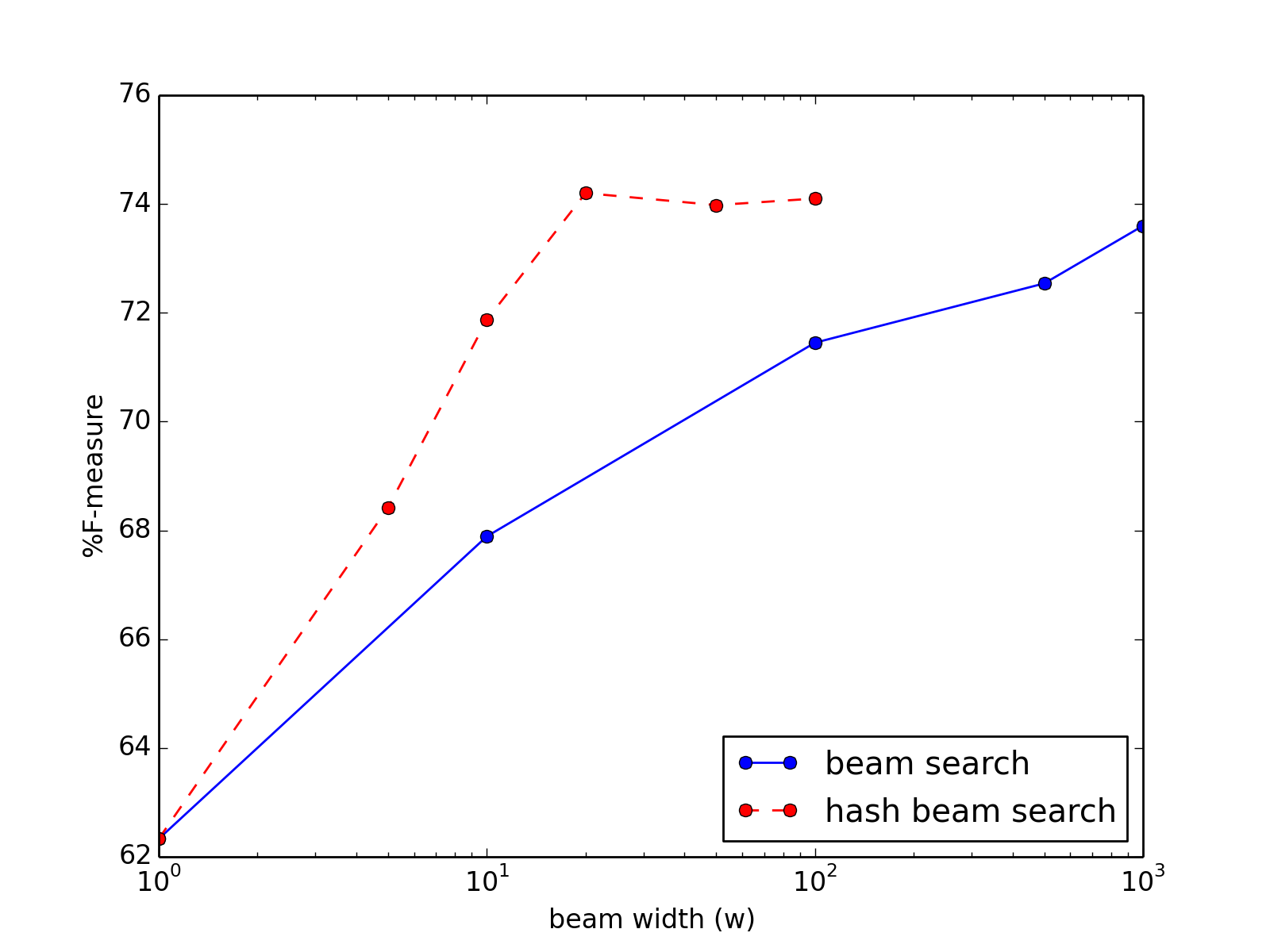}
\centering
\caption{Effect of beam width ($w$) on F-measure using evaluation Configuration 1. $k=2,K=4,f_h=y_t$.}
\label{beam_search}
\end{figure}

In Table \ref{f-scores}, we present F-scores (both frame and note based) for all the acoustic models and the 3 post-processing methods using Configuration 1. From the table, we note that all the neural network models outperform the PLCA and NMF models in terms of frame-based F-measure by $3\%-9\%$. The DNN and RNN acoustic model performance is similar, while the ConvNet acoustic model clearly outperforms all the other models. The ConvNets yield an absolute improvement of $\sim 5\%$ over the other neural network models, while outperforming the spectrogram factorisation models by $\sim 10\%$ in frame-wise F-measure. For the note-based F-measure, the RNN and ConvNet models perform better than the DNN acoustic model. This is largely due to the fact that these models include context information in their inputs, which implicitly smooths the output predictions. 

We compare the different post-processing methods for Configuration 1 by observing the rows of Table \ref{f-scores}. We note that the MLM leads to improved performance on both frame-based and note-based F-measure for all the acoustic models. The performance increase is larger on the note-based F-measure. The relative improvement in performance is maximum for the DNN acoustic model, compared to the RNN and the ConvNet. This could be due to the fact that the independence assumption in Equation \ref{independence} is violated by the RNN and ConvNet, which include context information while making predictions. This leads to some factors being counted twice and we observe a smaller performance improvement in this case. 
From Rows $1$ and $2$ of Table \ref{f-scores} we observe that the RNN-NADE MLM yields a performance increase for the PLCA and NMF acoustic models, though the relative improvement is less as compared to the neural network acoustic models. This might be due to the fact that unlike the neural network models, these models are not trained to maximise the conditional probability of output pitches given the acoustic inputs. Another contributing factor is the fact that the PLCA and NMF posteriograms represent the energy distribution over pitches rather than explicit pitch probabilities, which results in many activations being greater than $1$. 
This discrepancy in the \emph{scale} of the acoustic and language predictions leads to an unequal weighting of predictions when used in the hybrid RNN framework. In Table \ref{f-scores} we observe that the acoustic model in \cite{vincent2010adaptive} outperforms all other acoustic models on the note-based F-measure, while the frame based F-measure is significantly lower. This can be attributed to the use of an ERB filterbank input representation, which offers improved temporal resolution over the CQT for lower frequencies. 
\begin{figure*}[htpb!]
\centering
\subfloat[ConvNet Posteriogram]{\includegraphics[width=0.5\textwidth,height=7cm]{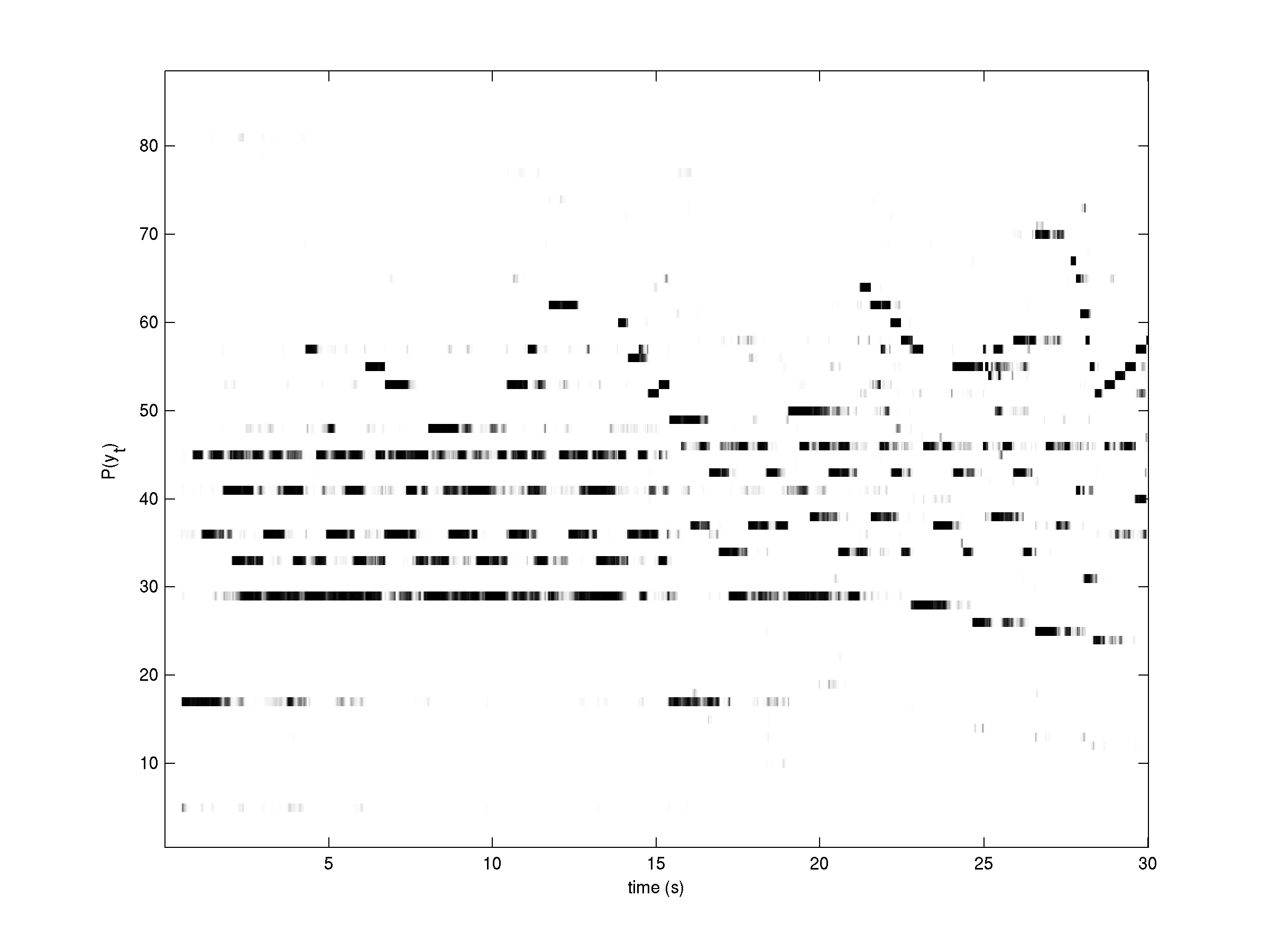}}
\subfloat[ConvNet Transcription]{\includegraphics[width=0.5\textwidth,height=7cm]{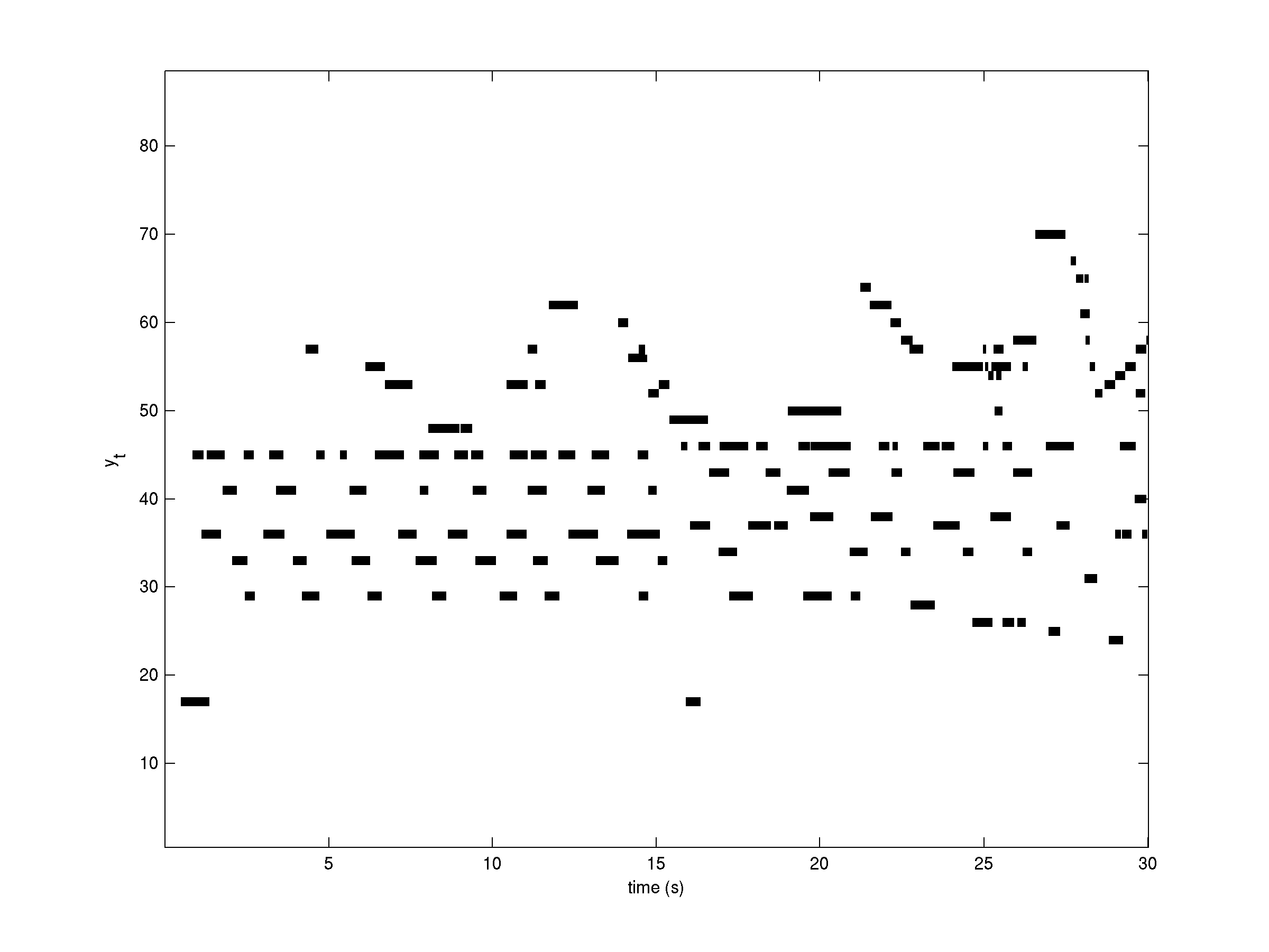}}\\
\subfloat[Ground Truth]{\includegraphics[width=0.75\textwidth,height=7.4cm]{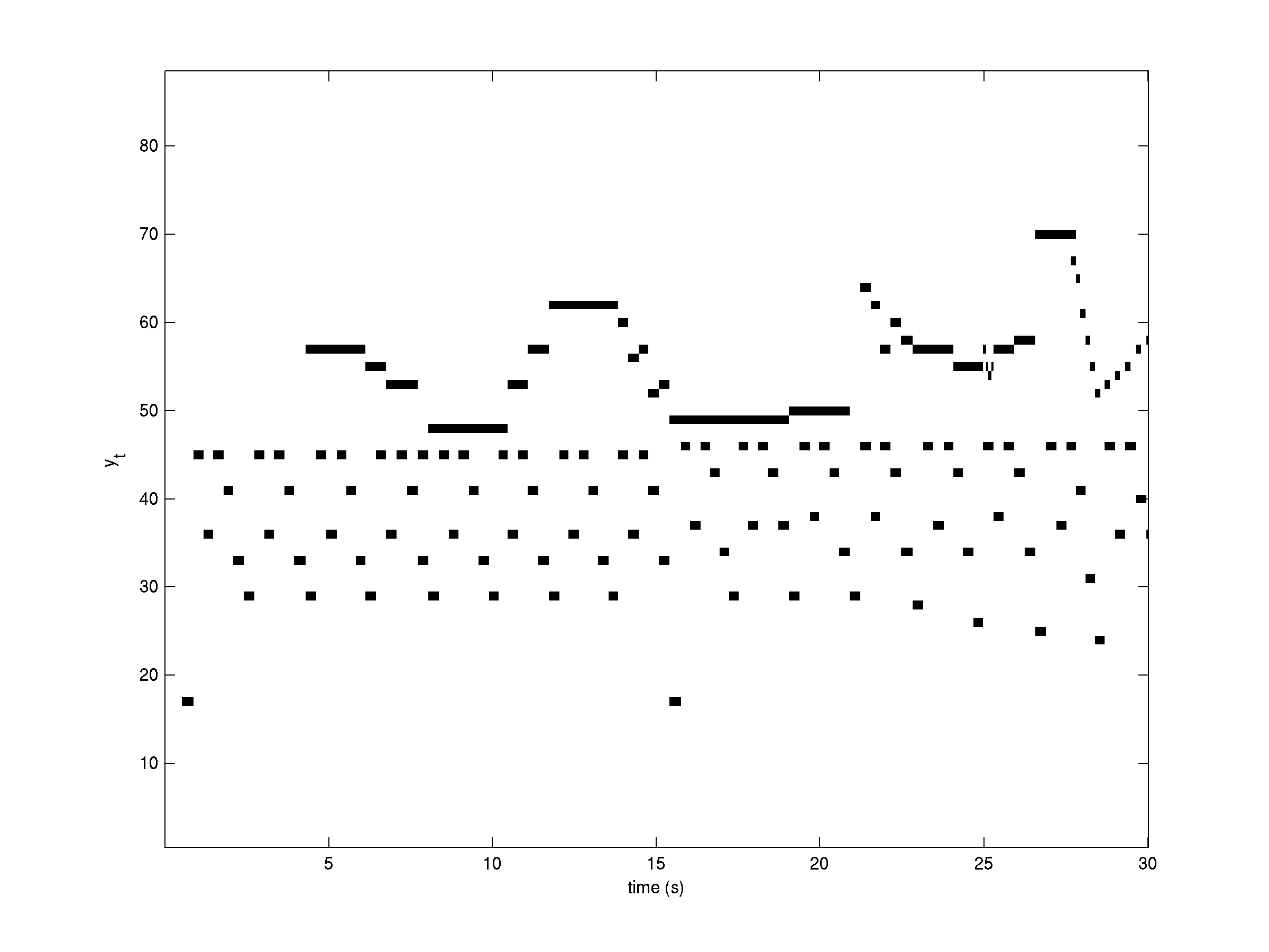}}\\
\caption{a) Pitch-activation (posteriogram) matrix for the first 30 seconds of track MAPS\_MUS-chpn\_op27\_2\_AkPnStgb produced by a ConvNet acoustic model. b) Binary piano-roll transcription obtained from posteriogram in a) after post processing with RNN MLM and beam search. c) Corresponding ground truth piano roll representation.}
\label{outputs}
\end{figure*}

In Table \ref{more-metrics}, we present additional metrics (precision, recall and accuracy) for the all the acoustic models after decoding with an RNN-MLM, using Configuration 1. We observe that that the NMF and PLCA models have low frame-based precision and high recall and the converse for the note-based precision. For the neural network models, we observe smaller differences between the both frame-based and note-based precision and recall values. Amongst all the neural network models, we observe that the ConvNet outperforms all the other models on all the metrics. 

In Table \ref{Disklavier}, we present F-measures for experiments where the acoustic models are trained on synthesised data and tested on real data (Configuration 2). From the table we note that frame based F-measure for the DNN and RNN models is similar to the PLCA model and the model in \cite{vincent2010adaptive}. We note that the ConvNet outperforms all other models on the frame-based F-measure by $\sim 5\%$. On the note based evaluations, we observe that both RNN and DNN are outperformed by all the other models. The ConvNet performance is similar to the PLCA model, while the acoustic model from \cite{vincent2010adaptive} again has best performance on the note based metrics. 

We now discuss details of the inference algorithm. The high dimensional hashed beam search algorithm has the following parameters: the beam width $w$, the branching factor $K$, number of entries per hash table entry $k$ and the similarity metric $f_h$ (Algorithm $2$). We observed that a value of $K \geq 4$ produced good results. Larger values of $K$ do not yield a significant performance increase and result in much longer run times, therefore we set $K= 4$ for all experiments. We observed that small values of $k$ (number of solutions per hash table entry), $1 \leq k \leq 4$ produced good results. Decoding accuracies deteriorate sharply for large values of $k$, as observed in \cite{sigtiachords}. 
Therefore, we set the number of entries per hash key  $k = 2$ for all experiments. We let the similarity metric be the last $n$ emitted symbols, $f_h(y_0^t) = y_{t-n}^t$. We experimented with varying the values of $n$ and observed that we were able to achieve good performance for small $n$, $1 \leq n \leq 5$. 
We did not observe any performance improvement for large $n$, therefore for all experiments we fix $f_h(y_0^t) = y_t$. Figure \ref{beam_search} is a plot showing the effect of beam width $w$ on transcription performance. The results are average values of decoding accuracies over $4$ splits. 
We compare performance of the hashed beam search with the high dimensional beam search in \cite{sigtia2014hybrid}. From Figure \ref{beam_search} we observe that the hashed beam search algorithm is able to achieve performance improvement with significantly smaller beam-widths. For instance, the high dimensional beam search algorithm takes $20$ hours to decode the entire test set with $w = 100$, while the hashed beam search takes $22$ minutes, with $w = 10$ and achieves better decoding accuracy. 

Figure \ref{outputs} is a graphical representation of the outputs of a ConvNet acoustic model. We observe that some of the longer notes are fragmented and the offsets are estimated incorrectly. One reason for this is that the ground truth offsets don't necessarily correspond to the offset in the acoustic signal (due to effects of the sustain pedal), implying noisy offsets in the ground truth. We also observe that the model does not make many harmonic errors in its predictions.

\section{Conclusions and Future Work}
\label{section:discussion}

In this paper, we present a hybrid RNN model for polyphonic AMT of piano music. The model comprises a neural network acoustic model and an RNN based music language model. We propose using a ConvNet for acoustic modelling, which to the best of the authors' knowledge, has not been attempted before for AMT. Our experiments on the MAPS dataset demonstrate that the neural network acoustic models, especially the ConvNet, outperform 2 popular acoustic models from the AMT literature. We also observe that the RNN MLMs consistently improve performance on all evaluation metrics. The proposed inference algorithm with the hash beam search is able to yield good decoding accuracies with significantly shorter run times, making the model suitable for real-time applications. 


We now discuss some of the limitations of the proposed model. As discussed earlier, one of the main contributing factors to the success of deep neural networks has been the availability of very large datasets. However datasets available for AMT research are considerably smaller than datasets available in speech, computer vision and natural language processing (NLP). Therefore the applicability of deep neural networks for acoustic modelling is limited to datasets with large amounts of labelled data, which is not common in AMT (at least in non-piano music). 
Although the neural network acoustic models perform competitively, their performance could be further improved in many ways. Noise or deformations can be added to training examples to encourage the classifiers to be invariant to commonly encountered input transformations. Additionally, the CQT input representation can be replaced by a representation with higher temporal resolution (like the ERB or a variable-Q transform), to improve performance on note based metrics.

The abundance of musical score data and recent progress in NLP tasks with neural networks provide strong motivation for further investigations into MLMs for AMT. Although our results demonstrate some improvement in transcription performance with MLMs, there are several limitations and open questions that remain. The MLMs are trained on binary vectors sampled from the MIDI ground truth. Depending on the sampling rate, most note events are repeated many times in this representation. The MLMs are trained to predict the next frame of notes, given an input sequence of binary note combinations. In cases where the same notes are repeated many times, log-likelihood can be trivially maximised by repeating previous inputs. This causes the MLM to perform a smoothing operation, rather than imposing any kind of musical structure on the outputs. A potential solution would be to perform beat-aligned language modelling for the training and the test data, rather than sampling the MIDI at some arbitrary sampling rate. 
Additionally, RNNs can be extended to include duration models for each of their pitch outputs, similar to second order HMMs. However, this is a challenging problem and currently remains unexplored. It would also be interesting to encourage RNNs to learn longer temporal note patterns by interfacing RNN \emph{controllers} with external memory units \cite{grefenstette2015learning} and also to incorporate a notion of timing or metre in the input representation for the MLMs. 

The effect of tonality on the performance of the MLMs should be further investigated. The MLMs should ideally be invariant to transpositions of a musical piece to different pitches. The MIDI ground truth can be easily transposed to any tonality. MLMs can be trained on inputs with transposed tonalities or individual MLMs for each key can be trained. Additionally, the fully connected input layer of the RNN MLM can be substitued with a convolutive layer, with convolutions along the pitch axis to encourage the network to be invariant to pitch transpositions. 

Another limitation of the proposed hybrid model is that the conditional probability in Equation \ref{independence} is derived by assuming that the predictions at time $t$ are only a function of the input at $t$ and independent of all other inputs and outputs. The violation of this assumption leads to certain factors being counted twice and therefore reduces the impact of the MLMs. The results clearly demonstrate that improvements with the MLM are maximum when the acoustic model is frame-based. The improvements are comparatively lower when combined with predictions from an RNN or ConvNet acoustic model. This is problematic since the ConvNet acoustic models yield the best performance.

\ifCLASSOPTIONcaptionsoff
  \newpage
\fi



%


\bibliography{biblio}
\bibliographystyle{IEEEtran}
%








\end{document}